# A Survey on Anomaly Detection for Technical Systems using LSTM Networks

**Benjamin Lindemann \*, Benjamin Maschler \*, Nada Sahlab \*, Michael Weyrich**

University of Stuttgart, Institute of Industrial Automation and Software Engineering, Pfaffenwaldring 47, 70569 Stuttgart, Germany, <first name>.<last name>@ias.uni-stuttgart.de

*These authors contributed equally to this publication.

**ABSTRACT** Anomalies represent deviations from the intended system operation and can lead to decreased efficiency as well as partial or complete system failure. As the causes of anomalies are often unknown due to complex system dynamics, efficient anomaly detection is necessary. Conventional detection approaches rely on statistical and time-invariant methods that fail to address the complex and dynamic nature of anomalies. With advances in artificial intelligence and increasing importance for anomaly detection and prevention in various domains, artificial neural network approaches enable the detection of more complex anomaly types while considering temporal and contextual characteristics. In this article, a survey on state-of-the-art anomaly detection using deep neural and especially long short-term memory networks is conducted. The investigated approaches are evaluated based on the application scenario, data and anomaly types as well as further metrics. To highlight the potential of upcoming anomaly detection techniques, graph-based and transfer learning approaches are also included in the survey, enabling the analysis of heterogeneous data as well as compensating for its shortage and improving the handling of dynamic processes.

**Keywords** Anomaly Detection, Artificial Intelligence, Autoencoder, Context Modeling, Long Short-Term Memory, Transfer Learning

## 1. INTRODUCTION

Anomalies pose a problem in various application areas, such as manufacturing, medical or communication systems. They often lead to a decrease in system performance and can cause instabilities and failure. Often, the causes of anomalies are unknown effects within complex systems. Hence, the capability of understanding and detecting these underlying effects with the aid of data is the key to ensure the desired outcome of complex technical systems. Due to the research progress in the field of machine learning, a wide range of new approaches for anomaly detection has been proposed in recent years. Different architectures of deep neural networks, and in particular, architectures based on Long Short-Term Memory (LSTM), have been designed proving to be capable of solving a variety of complex detection tasks, as is described in [1] or [2].

Existing high-profiled state-of-the-art surveys on anomaly detection techniques, such as [3] and [4], only marginally consider neural-network-based approaches. Current research developments regarding deep neural networks and LSTM architectures for anomaly detection are oftentimes not incorporated. The surveys mainly distinguish between statistical, classification-based, clustering-based and information-theoretic approaches. Thus, techniques based on the principal component analysis (PCA), the support vector machine (SVM), the k-nearest-neighbor (k-NN) algorithm or different types of correlation analysis constitute a major part of the investigations. The common ground for all approaches is their aim to detect anomalies based on static and time-invariant models [3, 4]. For the detection of dynamic and time-variant anomalies, additional techniques such as sliding windows are utilized and combined with the aforementioned approaches. In consequence, the approaches do not include models to adequately capture time-variant system dynamics and therefore cannot characterize anomalous contexts. To tackle the problem of detecting complex contextual anomalies with dynamic and time-variant characteristics, new promising recurrent neural network (RNN) architectures emerged. Survey studies regarding such deep learning approaches for anomaly detection have been conducted in recent years: In [5], advantages and disadvantages as well as computational complexity of (semi-)supervised, unsupervised and hybrid deep learning approaches are described. However, a classification of different LSTM approaches is missing and a detailed investigation of LSTM model architectures, scenario descriptions or detection mechanisms is not part of the work. An architectural analysis of deep-learning-based anomaly detection approaches with a focus on Boltzmann machines, Autoencoders (AE) and RNN is given by [6]. The survey conducted in [7] classifies existing detection approaches from the viewpoint of time series characteristics, namely approaches for uni- and multivariate time series data. LSTM approaches are mentioned in all of the existing deep learning surveys, but they are neither in central focus nor further classified and analyzed regarding architecture and detection mechanism. However, due to the recent emergence of different LSTM approaches that are widely used for different anomaly detection purposes, the present paper aims to present a detailed overview on anomaly detection for technical systems with a clear focus on such LSTM approaches.

The objective of this article is to give an overview of promising LSTM based approaches for anomaly detection with an additional focus on upcoming graph-based and transfer learning approaches. All approaches are evaluated based on a set of application-oriented criteria such as the detection capabilities regarding temporal anomalies, achieved accuracies and use cases addressed in the original publication.

This article is organized as follows: Chapter 2 introduces different anomaly types such as point, collective and contextual anomalies and gives an overview on temporal context modeling. Chapter 3 presents the investigated deep neural network and LSTM approaches for anomaly detection. The chapter is sectioned into regular LSTM as well as encoder-decoder-based and hybrid approaches. Chapter 4 then gives an overview on recent trends in deep-learning-based anomaly detection using graph-based and transfer learning approaches. Chapter 5 discusses the results of this survey. Chapter 6 concludes this article and points out new research directions.

## 2. ANOMALY CLASSIFICATION

Anomalies occur in various domains and are therefore subject to intensive research in a wide range of different application areas, such as in network security [4], internet of things [7], medicine [8] or manufacturing systems [9]. The essential common ground for all areas is the understanding of an anomaly as a deviation from the rule or of an irregularity that is not considered as a part of the normal system behavior [3, 5]. This matches with the definition in [8] where anomalies are described as abnormalities, deviants or outliers. Anomalous dynamics are mostly unknown and occur inadvertently, lead to instabilities and are therefore drivers of increased inefficiencies and system errors.

The taxonomy of anomalies applies for all investigated application areas: It can be characterized based on various aspects such as focus point (e.g. a certain actuator of a production machine), measurability, (non-)linearity and temporal behavior. Depending on the application, different focus points are possible. It can imply e.g. a direct affectation of the whole system's dynamics or an interference on the level of (individual) sensors and actuators. Anomalies can either be directly measurable or they have to be observed using some kind of indirect state estimation. Furthermore, they can either show linear or nonlinear characteristics. In the scope of the investigated literature, anomalous system dynamics being detected and modelled using LSTM networks are rarely time-invariant due to the properties of the LSTM cell. Regarding time-variant dynamics, it can be distinguished between stationary and non-stationary [10] or short-term and long-term behavior [11].

For the detection of such irregularities, originally, the main research focus was on stochastic methods for the detection of **outliers**, or so-called **point anomalies**. Thus, probability densities have been calculated for target parameters and defined percentiles have been declared as outliers and thereby as anomalies of a certain degree [12].

The overview of stochastic methods for anomaly detection presented in [3] introduces further types of anomalies that occur in time series data of technical systems. In consequence, collective and contextual disturbances are defined as further anomaly types in addition to the statistically described outliers. Fig. 1 illustrates the three anomaly types being incorporated into a univariate time series according to [13].

**Collective anomalies** can be characterized as a group of data vectors where each individual data vector is in tolerance, but the composition of the group indicates an irregularity. Hence, the internal structure of the data sequence determines the degree of deviation. In [14], collective anomalies are defined for the case of multivariate time series. A multidimensional metric based on the degree of deviation of all single data vectors of the group is proposed for their detection. Their significance is modeled in a time-variant manner with an exponentially decaying function. If the collective metric based on all single deviations exceeds a dynamic threshold, a collective anomaly is detected.

**Contextual anomalies** can be characterized as individual data vectors (being no point anomalies) or groups of data vectors (being no collective anomalies) that are in tolerance but indicate an irregularity in the scope of the specific surrounding data vectors or groups of data vectors, hereby referred to as context.

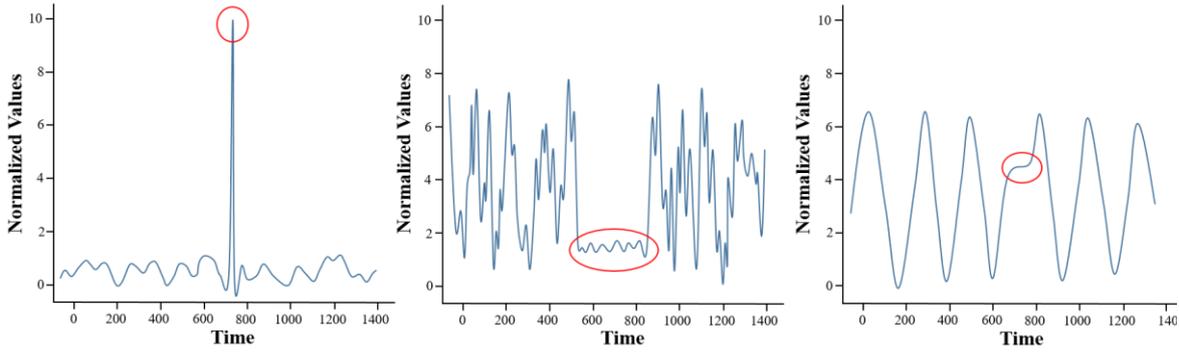

FIGURE 1. **Point anomaly (left), collective anomaly (middle) and contextual anomaly (right) according to [13]**

Applying contextual anomaly detection can contribute to a better anomaly detection by emphasizing or discarding a certain anomaly, thereby reducing false positives or by finding the root cause for a certain anomaly with the help of meta data, also referred to as anomaly attribution [15,16,17].

Therefore, in contrast to collective anomalies that are described by their internal structures or content, the detection of contextual anomalies highly depends on the short-term and long-term characteristics of the surrounding external data structures. With regard to multivariate time series, a context of a data vector or a group of data vectors can be interpreted as a union of all surrounding data vectors that lay in a defined time horizon. Distance-based metrics are primarily utilized for the characterization of contextual anomalies. This can be realized, for instance, based on a sliding window technique, of which the distance metric to a previous window is recalculated with every new data sample. The excess of dynamic thresholds indicates contextual anomalies [18]. Otherwise, the database of known contexts is further expanded. Depending on the application domain, context can have different definitions, scopes and dimensions. It can for example refer to temporal, spatial or spatio-temporal as well as further relevant environmental attributes. For the domain of discrete manufacturing, it can encompass data about the manufacturing process and the associated process parameters.

Contextual anomaly detection represents a twofold challenge: Firstly, a shared context has to be defined and detected and, secondly, anomalous data points need to be identified within it. Therefore, contextual – sometimes referred to as external – attributes and behavioral attributes have to be identified [16].

## 3. ANOMALY DETECTION WITH LSTM NETWORKS

This chapter begins with a short introduction of the LSTM cell to provide a basic understanding for network architectures and detection mechanisms of the LSTM networks discussed in the subsequent chapters. The LSTM cell has been developed by [19] to tackle the vanishing-gradient-problem that occurs with conventional RNN and leads to the inability to learn long-term dependencies. The vanishing-gradient-problem describes the circumstance where parts of weights in RNN tend to stop changing during the learning process. In consequence, a prioritization of current information could lead to neglecting past events. Thus, relations that recur over a long period of time (long-term dependencies) cannot be adequately learned. LSTM is constructed to control the whole information flow within neurons. For this purpose, a gating mechanism is introduced that controls the process of adding and deleting information from an iteratively propagated cell state. Thus, the process of forgetting can be controlled, and a defined memory behavior is realized to model both, short-term as well as long-term dependencies. The LSTM cell architecture is depicted in Fig. 2.

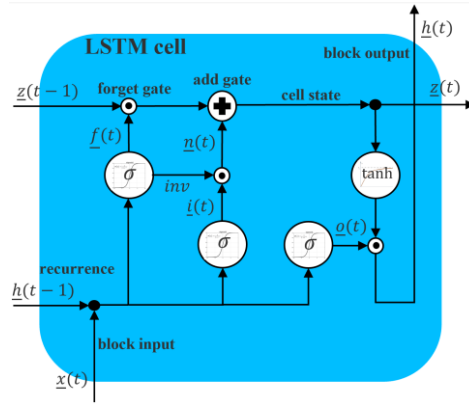

In contrast to conventional RNN, the output of the neuron $\underline{h}(t)$ is not directly constructed with inputs $\underline{x}(t)$ and previous outputs $\underline{h}(t-1)$, but based on the cell state $\underline{z}(t)$. The cell state on the other hand is determined by the LSTM gating mechanism. Vector $\underline{f}(t)$ as the output of the forget gate and vector $\underline{n}(t)$ as the output of the add gate iteratively adapt the cell state to control the memory behavior. In the depicted LSTM architecture, an inverse connection between these two gates is used to limit the memory capacity to a certain degree. Hence, with every iteration, information is added and deleted from the cell state. This procedure is driven by the circumstance that no memory is infinite and that the human memory as a role model also possesses a limited capacity. The output gate further infers the updated cell state to calculate the desired output. Therefore, the most recent inputs do not necessarily dominate the generation of the output signals, because the cell state encapsulates a reduced and weighted representation of historic input information. This information is mapped onto the output. Hence, the influence of e.g. important past events is incorporated in the projection of the output and the negligence of current inputs with low information density is feasible. The gates themselves are constructed based on current inputs and past outputs.

FIGURE 2. **LSTM cell with inverse mapping between forget and add gate according to [19]**

The LSTM cell can be incorporated into a wide range of neural network architectures. An overview of researched approaches is given in Table 1. In the further course of this article, we focus on the investigation of network architectures that are based on LSTM cells and that are developed to solve defined anomaly detection tasks. The subsequent chapter is divided into encoder-decoder-based, hybrid, graph-based and transfer learning approaches.

### 3.1 LSTM-BASED APPROACHES

LSTM networks are predestined to detect contextual anomalies due to their ability to learn temporal relations and to capture them in a low-dimensional state representation. Those relations can concern stationary and non-stationary dynamics as well as short-term and long-term dependencies. LSTM networks are particularly suitable for modeling multivariate time series and time-variant systems [20]. Hence, the deviation of real system outputs from expected outputs being predicted by the network can be utilized for anomaly detection purposes. LSTM-based approaches have proven to show excellent anomaly detection capability, such as in the fundamental work of [21]. The paper presents a stacked LSTM architecture to detect anomalies within time series data. In contrast to robust or denoising LSTM AE, no dimensionally reduced features are utilized as inputs. The detection is realized by evaluating the deviation of predicted outputs based on a variance analysis. In [22], a deep LSTM network is used as predictor of the regular bus communication behavior in vehicles. Significant deviations are detected using a dynamic threshold to detect anomalous communication behavior caused by cyber-attacks. In [23], a compound architecture is presented. Here, the LSTM network predicts regular system dynamics and a support vector machine is applied as classifier for anomalies to realize an adaptable and self-learning detection mechanism. Thus, temporal anomalies in multivariate data can be detected in a semi- or unsupervised manner. An approach to detect collective anomalies with LSTM networks is presented in [24]. The novelty consists of an evaluation of multiple one-step ahead prediction errors in contrast to evaluating each time step separately. LSTM networks enhance the detection accuracy by predictively modeling stationary and non-stationary time dependencies. Thereby, an efficient detection of temporal anomaly structures is realized. In [25] a real-time detection approach is realized based on two LSTM networks. One to model short-term characteristics and is able to detect single upcoming anomalous data points within time series and the other to control the detection based on long-term thresholds.

### 3.2 ENCODER-DECODER-BASED APPROACHES

In the majority of uses cases investigated in various application fields, such as manufacturing or communication, acquired data neither possesses any labels, nor are information models describing the data context available. In consequence, unsupervised learning methods are necessary to realize an indirect labeling of the data and to thereby detect anomalies. In particular, novel neural network approaches with an encoder-decoder architecture that have been developed in recent years show an excellent applicability for unsupervised detection tasks. AE networks are an example where the encoder part aims to learn a lower dimensional representation of the input data and the decoder part targets a reconstruction of these compressed features [26]. Hence, the AE is trained with data that represents normal system dynamics and learns how to compress and reconstruct this data. In contrast, the processing of anomalous data with the trained AE results in a reconstruction error. The error dynamics can be utilized to generate an anomaly detection mechanism, as presented in [27] for the example of a robust deep AE. In this case, a principal component analysis and regularization layers have been integrated in the AE to denoise the input data and to realize a robust detection behavior. To efficiently extract anomalous dynamics, the reconstruction metric consists of two parts. A similar approach is pursued with the contractive LSTM AE described in [28]. One part of the reconstruction metric evaluates the ability to separate anomalies such as outliers from the normal data and the other one evaluates the ability to discover relations within the data. In addition, [28] describes a denoising LSTM AE that aims to optimize prediction and detection accuracy by extracting underlying and uncorrupted relations in disturbed data.

**TABLE 1.** Overview of surveyed regular LSTM, encoder-decoder-based and hybrid approaches

| | Input | | | Model | | | Evaluation | | |
|---|---|---|---|---|---|---|---|---|---|
| Source | Data Type | Labels | Features Extracted | Anomaly Type(s) | Architecture | Adaptiveness | Scenario | Metrics | Performance |
| Malhotra et al. (2015) [21] | Multivariate time series | No | No | Contextual | Stacked LSTM | No | Several, such as power demand | Precision, recall, F1-score | Higher than RNN |
| Ergen et al. (2017) [23] | Multivariate time series | Both | No | Collective, contextual | Stacked LSTM-SVM | No | Several, e.g. HTTP requests | AUC, ROC, Sign functions | Higher than SVM and SVDD |
| Bontemps et al. (2016) [24] | Univariate time series | No | No | Collective | LSTM | No | Intrusion detection for PC networks | Average relative error, danger coefficient | No comparison with other methods carried out |
| Lee et al. (2020) [25] | Univariate time series | No | No | Outlier, collective | Dual LSTM | Yes | Multiple domain streaming data | Precision, recall, F1-score | Higher than competitors, real-time-capable |
| Zhou et al. (2017) [27] | Image sequences | Yes | Principal components | Outlier | Robust deep LSTM AE | Yes | Image processing | L1-norm | High on benchmark datasets |
| Naseer et al. (2018) [28] | Multivariate time series | Yes | Median and interquartile range | Outlier, collective, contextual | Contractive LSTM AE | No | Intrusion detection for computer networks | AUC, ROC, precision, recall, accuracy | Higher than conventional ML methods (ELM, k-NN, RF, SVM) |
| Park et al. (2018) [29] | Multivariate time series | No | No | Outlier, contextual | Variational LSTM AE | Yes | Environ. anomalies in robot systems | AUC, ROC, recons. error | Higher than HMM, SVM, AE |
| Lindemann et. al. (2020) [30] | Multivariate time series | No | Yes | Outlier, collective, contextual | Observer-based LSTM AE | Yes | Discrete manufacturing | Recons. error | No comparison with other methods carried out |
| Fernando et al. (2017) [32] | Image sequence | No | No | Outlier, collective | LSTM-AE | No | Pedestrian trajectory prediction | Spatial error metrics | No comparison with other methods carried out |
| Loganathan et al. (2018) [33] | Univariate time series | No | No | Outlier | Seq2Seq LSTM | No | Intrusion detection for computer networks | Accuracy, sequence length | Higher than competitor |
| Zenati et al. (2018) [37] | Several | No | Scaling, one-hot representation | Outlier, collective, contextual | GAN + LSTM | No | Several, such as image processing | Precision, recall, F1-score | Higher than VAE, SVM |
| Kim et al. (2018) [38] | Multivariate time series | No | Based on convolution | Outlier, collective, contextual | CNN + LSTM | No | Web traffic | Cross entropy | Higher than RF, MLP, KNN |
| Ding et al. (2020) [39] | Multivariate time series | No | No | Contextual | LSTM + EWMA | No | Industrial robotic manipulators | Precision, recall | No comparison with other methods carried out |

**Legend:** *'Data Type'* refers to the type and dimensionality of input data, e.g. multi-variate time-series or RGB images. *'Labels'* details whether the input data is labeled, e.g. as 'normal' or 'anomalous'. *'Features Extracted'* details whether the input data has been subject to a feature extraction process and with what method such a process would have been carried out. *'Anomaly Type(s)'* refers to the type(s) of anomalies an algorithm can detect. See chapter 2 for details on those types. *'Architecture'* refers to an algorithm's network architecture, e.g. the type of cells used. *'Adaptiveness'* refers to an algorithm's capability to expand its knowledge during operations, e.g. to detect previously unknown anomalies. *'Scenario'* details the use case on which an algorithm's performance is evaluated. *'Metrics'* details the metrics used for evaluating an algorithm's performance. Table 2 lists the most significant metrics used in the following chapter's approach evaluations. *'Performance'* details an algorithm's performance compared to other approaches evaluated on the evaluation scenario.

**TABLE 2.** Detection approach performance metrics

| Metric | Definition |
|---|---|
| Precision | *Precision* measures the number of actual anomalies being detected in relation to all detected anomalies. |
| Recall | *Recall* measures the number of detected anomalies in relation to all actual anomalies. |
| Accuracy | *Accuracy* measures the number of actual anomalies being detected and normal data instances being classified as such in relation to the entire data set |
| F1-Score | *F1-Score* is calculated based on precision and recall and measures the quantity of any type of false detections in detection mechanisms. |
| Receiver operating characteristic curve (ROC) | *ROC* describes a curve that visualizes the ratio of correctly detected anomalies against incorrectly detected anomalies for varying thresholds. |
| Area under the curve (AUC) | *AUC* Is the integral under the ROC. A high value represents a model with high recall and low false positive rate |
| Cross entropy | *Cross entropy* compares distributions in terms of a quantification of their difference. Hence, detected anomaly distributions can be evaluated based on known distributions of test data sets. |

The composition of LSTM and AE allows learning short-term and long-term dependencies in terms of temporal lower-dimensional features and thereby provides a found basis for detecting complex time-variant anomalies. Based on the described work, further extensions and studies have been conducted on anomaly detection with LSTM AE. In [29], a variational LSTM AE is introduced for anomaly detection purposes. The scheme utilizes probabilistic projection mechanisms in the encoder and decoder part. Thereby, input sequences are transformed to lower-dimensional feature distributions and reconstructed based on a defined feature value. The approach applies a log-likelihood-based anomaly detection by calculating a log-likelihood score for real and reconstructed outputs. In addition, the trained encoder can be used separately for probabilistic dimension reduction purposes. Another approach is presented in [30] where an LSTM AE is used to model the normal system behavior of discrete manufacturing processes. The trained decoder part of the network is utilized as inverse process model to detect anomalies by a disturbance observer-based comparison of real and reconstructed actuating variables. Hence, effects of different characteristics, such as stationary and non-stationary anomalies, that disturb the actuation systems can be detected.

In addition to AE, sequence-to-sequence (Seq2Seq) LSTM networks possess an encoder-decoder structure and have been used for anomaly detection tasks [31]. This network type is utilized in [32] to detect anomalies based on the cell states being propagated through the network. Unknown cell states and highly deviating copying vectors between encoder and decoder layer are considered as anomalies. They are further evaluated by a postprocessing clustering algorithm. Another Seq2Seq approach is described in [33] where different anomaly types can be detected by modeling and predicting a variety of attributes. This approach has outperformed the stacked LSTM by [21] in the scope of an empirical study on a benchmark data set. An approach to enhance the generalization and extrapolation abilities of Seq2Seq LSTM networks for an optimized anomaly detection is presented by [34]. The proposed architecture is illustrated in Fig. 3. It consists of sparsely connected encoders and decoders containing skip connections that depend on the information density in the input sequences and enable a more flexible propagation of the cell state. Multiple encoders use the same copying layer to propagate a reduced feature vector to the decoder. This procedure prevents overfitting and leads to better generalization characteristics. The cost function for detecting anomalies minimizes the entirety of all reconstruction errors and contains a penalty term to control the information flow in the joint copying layer.

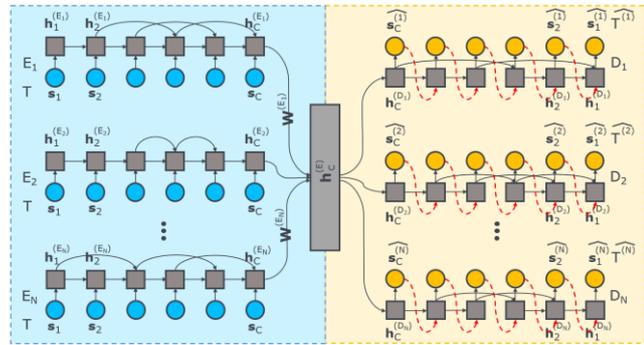

FIGURE 3. **Ensemble of LSTM AE with skip connections according to [34]**

### *3.3 HYBRID APPROACHES*

The approaches described in this chapter have the commonality of using a composition of two neural networks in the scope of a compound anomaly detection architecture. Thus, these hybrid approaches consist of an LSTM and a second network. Primarily, one copes with the task of predicting process dynamics and the second detects deviations from the actual process outcome and detects anomalous dynamics. The aim of hybrid approaches is to benefit from the advantages of both network types while simultaneously compensating inaccuracies. The occurrence of local anomalies consists of deviations with regard to single or multiple data instances and is temporarily restricted to a fix time span. On the contrary, global anomalies primarily describe drifting deviations from the reference that explicitly show non-stationary long-term characteristics. In [35] a novel composition of stacked AE and LSTM is presented for the detection of anomalies based on unlabeled data and unknown system dynamics. The encoder part is constructed to process multiple sequences at every discrete time step and can either process raw data or reduced input features. It extracts relations by maximizing the entropy within the compressed information. The anomaly detector is realized using a second network in the scope of a compound architecture. Hence, a LSTM network is trained to identify deviation characteristics in the reconstructed feature space. A further example for a hybrid approach with LSTM AE is presented in [36] where it is extended by a clustering algorithm that characterizes reconstructed system dynamics using a state space representation. Hence, anomalous dynamics are identified in the case of an abrupt or drifting state transition or in the case of the creation of new states. To further optimize the cooperation of predictor and detector, generative adversarial networks based on LSTM are utilized in [37]. The network architecture consists of two interacting networks, the generator that aims to replicate the data of the real system and the discriminator that targets to distinguish this artificially generated data from real data. The generator behaves like a decoder and reconstructs real time series data whereas the discriminator classifies the reconstruction as originated from normal or anomalous inputs. This depends on the difficulty of distinguishing the reconstruction from real data.

To enable a multidimensional anomaly detection, convolutional neural networks (CNN) and LSTM are combined in the method developed by [38]. The ability to efficiently compress high-dimensional data enables extracting dependencies in several dimensions. Thereby, the compound use of CNN and LSTM allows anomaly detection in multiple and interconnected dimensions such as spatial, temporal or other application-specific dimensions. Hence, it is possible to detect complex contextual anomaly structures by correlating different dimensions even if they do not show anomalous behavior in all dimensions. The classification is conducted based on the cross entropy. The approach is illustrated in Fig. 4. In the work of [39], LSTM is combined with Exponentially Weighted Moving Average (EWMA) and a dynamic thresholding technique. Temporal structures of multivariate time series are analyzed to extract patterns of similar characteristics. These patterns are continuously identified and evaluated. The EWMA is used in combination with the dynamic thresholding to evaluate the prediction results of the LSTM network. This allows the detection of anomalous structures in the data by investigating the prediction residuals with the aforementioned technique. A major advantage of the approach is the fact that all contextual anomalies of new time sequences can be identified in the scope of a single detection process so that a significant efficiency increase can be achieved.

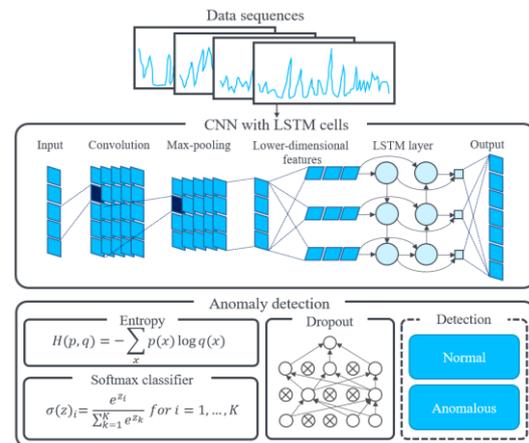

FIGURE 4. **CNN and LSTM for anomaly detection according to [38]**

## 4. RECENT TRENDS IN LEARNING-BASED ANOMALY DETECTION

In this chapter, recent trends in deep-learning-based anomaly detection are presented. They aim at expanding the versatility and robustness of the approaches described in chapter 3: Graph-based approaches allow for an improved representation of contextual information, whereas transfer learning approaches focus on the amount of data needed for training anomaly detection algorithms.

### *4.1 GRAPH-BASED APPROACHES*

A graph is a network of vertices and edges, which can be directed or undirected with weighted or unweighted edges depending on the nature and the domain of data being modeled. Detecting anomalies within a graph or using a graph-based approach represents further promising approaches for anomaly detection. The main advantage of these approaches is a graph's capability to model correlations between datapoints rather than individually representing them. This way, the inherent interdependencies between data, also with reference to other external factors can be highlighted and analyzed. Another advantage is the applicability of graphs for detecting collective and contextual anomalies by clustering nodes based on contextual attributes and detecting anomalous edges or nodes within the clusters. Graph-based anomaly detection has been present in research in the past decades, with mostly a focus on static graph analysis. With emerging machine learning and deep learning algorithms, dynamically evolving graphs over time are also considered for anomaly detection [15,16].

In essence, detecting anomalies within graphs requires two or three steps depending on the structure of input data. Fig. 5 depicts a simplified overview of the process. Initially, either data is already represented in a graph or needs to be modeled in one from various heterogeneous databases and data formats, where the graph schema with nodes and edges are defined depending on the use case. Graph models are then stored and managed within a graph database while being versioned based on different time instances to enable the analysis of time-based anomaly progression. A second step generally includes a clustering or partitioning of the graph to sub-graphs, which can be based on structural or temporal features as well as a combination of both. This initial graph analysis helps identify the network structure as well as weakly or strongly connected nodes and clusters. Furthermore, context and content features can be defined and detected. Building on this step, the third step requires a time-based analysis within the clusters to detect anomalous nodes or edges, based on the defined graph model and contextual features. To mark these clusters, node edges, i.e. relationships, can be iteratively assigned weights.

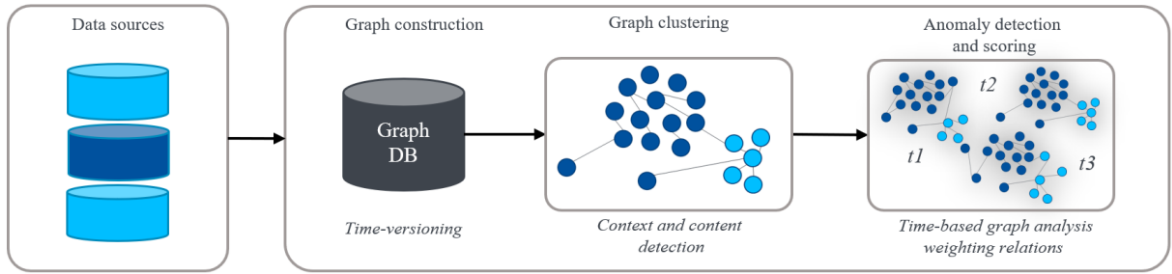

FIGURE 5. **Graph-based approach for anomaly detection as a pre-step to LSTM-based approaches**

The result of this described process can be further used and enhanced in different possible ways. The initially identified anomalous clusters can be further investigated by applying graph embeddings, i.e. transforming the sub-graph structure to a vector representation to apply additional machine learning algorithms. Alternatively, the resulted features can be used as an input for LSTM networks, therefore enabling a more efficient use of them.

Subsequently, some existing approaches in literature for graph-based anomaly detection will be presented. Primarily, the authors were interested in finding approaches for contextual anomaly detection for time-series data using graphs – preferably as a combination or extension for LSTM. Due to the lack of investigations of these approaches in literature, some representative approaches with varying complexity and scope have been chosen out, which can be seen on Table 3.

In [40], an algorithm for detecting contextual collective anomalies is presented. Having an attributed graph as an input, nodes represent individual datapoints, thus behavioral attributes and edges contextual attributes. Manual feature selection within the graph is suggested and a Louvain clustering algorithm used. Within the clusters, an anomaly score was given to each node in order to detect anomalous nodes.

In [41], contextual outliers in sensor data are detected in a graph-based approach. Behavioral attributes are given by temperature and humidity values, whereas the contextual attribute is represented by time. From a stored server dataset, a graph is created, representing each datapoint as a node and the edges based on the Euclidean distance and an assigned weight accordingly. Based on the graph, clusters are built with an iterative minimum spanning tree algorithm. Outliers are then detected using a sliding time window, where the cluster with the most data is considered as normal and other clusters as anomalous with a higher assigned outlier score. Training or labeling data are not necessary in this scenario.

Anomalies tend to have a dynamic nature and detecting them within dynamic graphs is a challenging task, because not only structural and content, but also temporal features should be considered. Those can be detected by capturing long-term as well as short-term node patterns.

In recent years, approaches for detecting anomalies within dynamic graphs have evolved, taking advantage of deep learning. Graph embeddings for instance map a graph to a vector space and analyze nodes based on their structural similarity. Furthermore, using graph convolutional networks (GCN) allows for an extension by extracting structural and content feature of nodes. Using GCN, a node's anomalous probabilities can be propagated to neighboring nodes. However, there still is little consideration regarding long-term feature detection. Extending on GCN to consider temporal features, Zheng et al [42] build upon gated recurrent unit (GRU) for anomaly detection

TABLE 3. Overview of surveyed graph-based approaches

| | *Input – Graph Type* | | | *Model* | | | *Evaluation* | | |
|---|---|---|---|---|---|---|---|---|---|
| Source | Data Type | Labels | Features Extracted | Anomaly Type(s) | Architecture | Adaptiveness | Scenario | Metrics | Performance |
| Prado-Romero *et al.* (2016) [40] | Attributed graph and attribute identifiers | Yes | Yes | Collective contextual | Graph-clustering and outlier score function | No | Amazon dataset of purchased products | AUC | Slightly higher than outlier ranking in clustered attributed graphs |
| Haque *et al.* (2018) [41] | Multi-variate time series – dynamic weighted graph | No | No | Contextual | Minimum-spanning tree clustering and voting scheme | Yes | Wireless sensor network | Average Accuracy, Precision Recall | No comparison with other methods carried out |
| Zheng *et al.* (2019) [42] | Time-stamped dynamic graph | No | Yes | Anomalous edges - outlier | GCN with GRU | No | Directed network of messages | AUC | Higher compared against 3 graph-outlier algorithms |

**Legend:** '*Data Type*' refers to the type and dimensionality of input data, e.g. multi-variate time-series or RGB images. '*Labels*' details whether the input data is labeled, e.g. as 'normal' or 'anomalous'. '*Features Extracted*' details whether the input data has been subject to a feature extraction process and with what method such a process would have been carried out. '*Anomaly Type(s)*' refers to the type(s) of anomalies an algorithm can detect. See chapter 2 for details on those types. '*Architecture*' refers to an algorithm's network architecture, e.g. the type of cells used. '*Adaptiveness*' refers to an algorithm's capability to expand its knowledge during operations, e.g. to detect previously unknown anomalies. '*Scenario*' details the use case on which an algorithm's performance is evaluated. '*Metrics*' details the metrics used for evaluating an algorithm's performance. Table 2 lists the most significant metrics used in the following chapter's approach evaluations. '*Performance*' details an algorithm's performance compared to other approaches evaluated on the evaluation scenario.

purposes. They propose an approach for detecting anomalous network edges by also using a GRU as an LSTM variant with a contextual attention-based model to capture short and long-term node similarities. The GCN therefore outputs a node state considering its structural and content features and is complemented with a GRU for long-term information capturing.

In conclusion, it can be stated that encompassing further available meta-data and external attributes when detecting anomalies can lead to better detection results or provide more insight for understanding the causes of anomalies. This effect increases when combined with graph-based methods. With data being represented in a graph, contextual information can be used to enhance anomaly detection. Detected and evaluated anomalies can then be an input to an LSTM, where the highly dynamic process behavior is modeled with uncertainties.

### *4.2 TRANSFER LEARNING APPROACHES*

Similar to other use cases, the widespread utilization of data-driven anomaly detection in manufacturing needs to overcome a major challenge: Datasets required for training such algorithms need to be large and diverse, making them hard to acquire [43, 44]. To overcome this challenge, transferring knowledge between several learning agents training on independent datasets can be a valuable approach.

An area of research focusing on utilizing knowledge acquired while training on previous tasks to improve the training of a new task is the field of transfer learning [45]. Here, different approaches have been identified to solve this kind of problem: a) transferring either instances (or transformations thereof) of the old tasks' datasets to the new one or b) (parts of) the algorithm itself (or transformations thereof). Although the former is usually easier to implement, it does not address the challenges named above. Therefore, only the latter shall be further considered. It consists of parameter transfer, a (partial) re-use of a network pre-trained on the source task, and relational knowledge transfer, which enhances parameter transfers with domain adaption. Both reduce the need for training data on the target task (see Fig. 6) [46].

Despite the practical application of transfer learning methods still being in its earliest stages, some exemplary implementations of anomaly detection algorithms can be found (see Table 4). Unfortunately, due to a lack of benchmark datasets, their performances are not directly comparable.

In [47], transfer learning is used to improve anomaly detection in electricity consumption across different aluminum extrusion machines. A denoising AE is pre-trained unsupervisedly on a large source dataset and then fine-tuned on the target dataset (see Fig. 6, middle). Unfortunately, no reference values acquired with other algorithms are given, so that the detection performance cannot be evaluated.

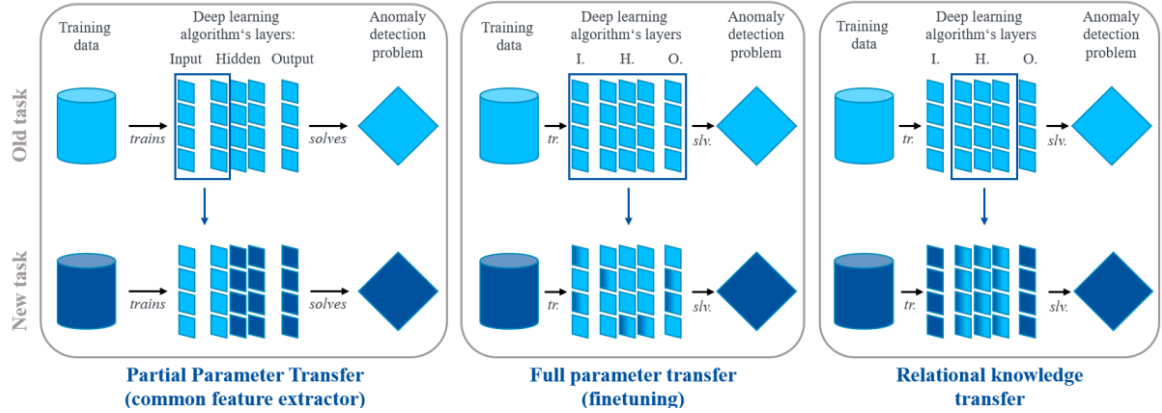

FIGURE 6. Transfer learning approaches: parameter transfer and relational knowledge transfer

In [48], transfer learning is used to improve anomaly detection in the operation of service elevators by allowing for differences in the type and number of sensors at different operating locations. This implementation relies on one-dimensional CNNs for each sensor to extract features. Those features are then aggregated and classified by different kinds of neural networks, among them LSTM, GRU and RNN (see Fig. 6, left). The ensuing analysis is methodologically very thorough, considering multiple distributions of 'normal' and 'anomalous' samples as well as different algorithms on this very elaborate application scenario. Although the accuracy achieved by the set of proposed algorithms is very high, unfortunately, no competing approaches are tried on the dataset.

In [49], transfer learning is used to improve anomaly detection on process sequences across different production modules. Here, the last encoding and decoding layer of a pretrained LSTM-based AE are re-trained on the respective target datasets. This allows for a swift adaption of a roughly pre-trained algorithm to the different production modules for which only small, normally insufficient datasets are available. Although showing better performances than conventional deep learning algorithms, the informative value is limited due to the simple process sequence dataset being far from realistic.

TABLE 4. Overview of surveyed transfer learning approaches

|  | *Input* | | | *Model* | | | *Evaluation* | | |
|---|---|---|---|---|---|---|---|---|---|
| **Source** | *Data Type* | *Labels* | *Features Extracted* | *Anomaly Type(s)* | *Architecture* | *Adaptiveness* | *Scenario* | *Metrics* | *Performance* |
| Liang et al. (2017) [47] | Uni-variate time series | No | No | Outlier | Denoising-AE | Yes | Electricity consumption | False rate, missing rate | No comparison with other methods carried out |
| Canizo et al. (2019) [48] | Multi-variate time series | Yes | No | Outlier, collective, contextual | CNN + varr. | Yes | Service elevator operation | Precision, recall, F1-score | No comparison with other methods carried out |
| Hsieh et al. (2019) [49] | Multi-variate time series | No | No | Collective, contextual | LSTM + AE | Yes | Simple production process | Precision, recall, F1-score | Higher than statistical methods or CNN |
| Tariq et al. (2020) [50] | Multi-variate time series | Yes | No | Collective, contextual | Conv-LSTM | Yes | Intrusion detection on CAN | Precision, recall, F1-score | Higher on unknown anomalies than SVM, IF and eRNN |
| Maschler et al. (2021) [51] | Uni-variate time series | Yes | No | Outlier, collective, contextual | LSTM | No | Discrete manu-facturing | Accuracy | Higher than non-transfer approach |

**Legend:** *'Data Type'* refers to the type and dimensionality of input data, e.g. multi-variate time-series or RGB images. *'Labels'* details whether the input data is labeled, e.g. as 'normal' or 'anomalous'. *'Features Extracted'* details whether the input data has been subject to a feature extraction process and with what method such a process would have been carried out. *'Anomaly Type(s)'* refers to the type(s) of anomalies an algorithm can detect. See chapter 2 for details on those types. *'Architecture'* refers to an algorithm's network architecture, e.g. the type of cells used. *'Adaptiveness'* refers to an algorithm's capability to expand its knowledge during operations, e.g. to detect previously unknown anomalies. *'Scenario'* details the use case on which an algorithm's performance is evaluated. *'Metrics'* details the metrics used for evaluating an algorithm's performance. Table 2 lists the most significant metrics used in the following chapter's approach evaluations. *'Performance'* details an algorithm's performance compared to other approaches evaluated on the evaluation scenario.

In [50], transfer learning is used to improve intrusion detection on controller area networks (CAN) by enabling the algorithm to learn new types of intrusion upon first encounter. Convolutional LSTM (ConvLSTM) are utilized to process the multi-variate time-series input as 2-D information, whereas Bayesian one-shot learning [51] provides transfer learning functionalities. An evaluation is carried out on CAN data collected from two different types of cars during more than 24 hours of driving. The algorithm outperforms several baseline models, e.g. SVM, isolation forests (IF) or ensemble RNNs (eRNN), on tasks that were not included in the training data, i.e. new tasks. On known tasks, it performs slightly worse than the best competing approach. Compared with the CAN data rate, the algorithm detects intrusions in real time.

An altogether different approach is used in [52]: Here, continual learning, i.e. multi-task machine learning using knowledge transfer between different tasks in order to benefit performance on old and new tasks, is used to improve anomaly detection. The use case is a metal forming process involving frequent changes of manufactured products. A stacked LSTM is combined with an enhanced loss function, causing the algorithm to retain previous capabilities. A thorough comparison of different continual learning approaches and a baseline conventional algorithm reveal significant improvements using so-called (online) elastic weight consolidation [53, 54].

Concludingly, there have only a few approaches been examined, covering merely a small subset of possible methods and techniques. While altogether promising, no clear trends can therefore be discerned yet, calling for more research in this area.

## 5. DISCUSSION

This paper gives an overview on LSTM networks for anomaly detection and divides existing approaches in five categories. The investigated, conventional neural network-based approaches are then further divided into regular LSTM architectures, encoder-decoder based LSTM networks as well as hybrid approaches.

Chapter 3 shows that regular LSTM allow the precise detection of collective and contextual anomalies. Compared with those pure LSTM approaches (see subchapter 3.1), encoder-decoder based architectures enable LSTM to further optimize their detection abilities for high-dimensional data spaces (see subchapter 3.2). Different architectures, such as contractive AE and variational AE, have been utilized for specific detection purposes. Hybrid approaches as described in subchapter 3.3 primarily aim to combine the benefits of two methods in one architecture. These approaches often contain a predictor and detector component so that the tasks are precisely divided.

Subchapter 4.1 outlines, that graph-based approaches for anomaly detection have the advantage of enabling a unified representation of heterogeneous data and data sources. This way, the cause and propagation of anomalies can be analyzed, especially within their contextual frame. Thus, many graph-based approaches address contextual anomalies. As an example, anomaly detection within physical processes can consider data about the process, the system as well as environmental attributes within a single graph to infer anomalies. Applying machine learning in combination with graph-based data representations and graph-analytics would theoretically be a promising approach for more accuracy in detecting and predicting anomalies especially in networked systems. However, so far, no actual implementations have been published.

Some open challenges remain regarding graph-based approaches, namely their input data structure. Compared with the challenge of lacking labeled data for machine learning, representing data within a graph and managing vertices and edges based on the domain application can be a complex and time-consuming task. A further challenge is the selection of contextual features within the graph and the complexity of encompassing multi-dimensional context features in the clustering and outlier ranking process. This problem is referred to as context-profiling, where multi-dimensional context cannot be discretized and requires a multi-variate graph-clustering approach. Finally, the lack of publicly available benchmark datasets hinders scientific advancement in this area as direct comparisons of different methodologies are hard to obtain.

As outlined in subchapter 4.2, transfer learning addresses the challenge of frequently not having sufficiently large and diverse datasets for training deep learning algorithms in anomaly detection use cases. By sequentially training on multiple datasets representing different tasks or different states of one task and transferring knowledge from one training to another, both challenges can be mitigated. Although still fairly new, there is a growing number of implementations reflecting the different approaches in conventional neural network-based learning (see chapter 3) being evaluated on industrial application use cases. It could be shown that transfer learning's capability to combine different datasets promises to allow training across systems and scenarios for a mutual benefit.

So far, to the authors' knowledge, there are no publicized results on combinations of transfer and graph-based learning. However, based on the presented findings, such a combination of approaches would be needed for the big challenges of practical machine learning facing the industrial community today: Detecting anomalies, e.g.

intrusions or potentially un-safe behavior in autonomous systems has hitherto proven to be too complex for conventional approaches. This is caused by these systems' high level of independence of pre-defined rules and their adaptability to new situations, constituting degrees of freedom that make it very hard to successfully discern between anomalies and normal, but newly learnt behavior.

## 6. CONCLUSION

Anomalies occur in a wide range of technical applications and can have significant effects on the performance and stability of the system as well as the quality of its output. They can be described by different characteristics. LSTM networks allow to particularly detect temporal characteristics. Hence, in the scope of the present work, different LSTM approaches for anomaly detection in time series data have been investigated regarding a detailed set of criteria. Additionally, recent advances in graph-based and transfer learning approaches towards anomaly detection were surveyed, focusing especially on their level of applicability towards real life problems.

Existing surveys on anomaly detection techniques deliver broad overviews on popular statistical, machine learning and deep learning approaches. However, they lack focus on current trends and the dynamic development that evolves in the area of neural networks for anomaly detection. Therefore, this paper presents anomaly detection approaches based on LSTM networks that have been applied in different technical systems, such as manufacturing or robotics. The conducted study indicates that different LSTM network architectures are available and capable of precisely detecting a varying range of complex anomalies, such as collective and contextual anomalies. Thereby, this article provides a range of state-of-the-art examples and analyses for anyone considering to enter the field of anomaly detection.

From the presented study, it is concluded that further research should be conducted regarding the detection of anomalies not solely in delimited systems but also in networks of interacting systems. Thus, future developments could focus on the incorporation of LSTM networks into graph-based approaches for an optimized characterization of contextual anomalies. To further enhance the detection accuracy, the combination of LSTM networks and transfer learning techniques could be examined more intensively to investigate the potential of transferring detected anomaly characteristics and knowledge between systems or networks of systems.

## REFERENCES

[1] G. Loganathan, J. Samarabandu, and X. Wang, "Sequence to Sequence Pattern Learning Algorithm for Real-Time Anomaly Detection in Network Traffic," IEEE Canadian Conference on Electrical & Computer Engineering (CCECE). IEEE, pp. 1–4, 2018. DOI: 10.1109/CCECE.2018.8447597

[2] H. Zenati, C. S. Foo, B. Lecouat, G. Manek, and V.R. Chandrasekhar, "Efficient GAN-Based Anomaly Detection," arXiv:1802.06222, 2018.

[3] V. Chandola, A. Banerjee, and V. Kumar, "Deep Learning for Anomaly Detection: A Survey," ACM Comput. Surv. 41 (3), pp.1–58, 2009. DOI: 10.1145/1541880.1541882

[4] M. Ahmed, A. Naser Mahmood, and J. Hu, "A survey of network anomaly detection techniques," Journal of Network and Computer Applications, Vol. 60, pp. 19–31, 2016. DOI: 10.1016/j.jnca.2015.11.016

[5] R. Chalapathy, and S. Chawla, "Deep Learning for Medical Anomaly Detection – A Survey," pp. 1-50, 2019. arXiv:1901.03407

[6] D. Kwon, H. Kim, J. Kim, S.C. Suh, I. Kim, and K.J. Kim, "A survey of deep learning-based network anomaly detection," Cluster Computing, Vol. 22, pp. 949-961, 2019. DOI: 10.1007/s10586-017-1117-8

[7] A.A. Cook, G. Mısırlı, and Z. Fan, "Anomaly Detection for IoT Time-Series Data: A Survey," IEEE Internet of Things Journal, Vol. 7, no. 7, pp. 6481 - 6494, 2019. DOI: 10.1109/JIOT.2019.2958185

[8] T. Fernando, H. Gammule, S. Denman, S. Sridharan, and C. Fookes, "Deep Learning for Medical Anomaly Detection – A Survey," pp. 1-20, 2020. arXiv:2012.02364

[9] G. A. Susto, M. Terzi, and A. Beghi, "Anomaly Detection Approaches for Semiconductor Manufacturing," Procedia Manufacturing, Vol. 11, pp. 2018-2024, 2017. DOI: 10.1016/j.promfg.2017.07.353

[10] V. Kuznetsov, and M. Mohri ,"Time series prediction and online learning," 29th Annual Conference on Learning Theory, Bd 49. PMLR, Columbia University, pp. 1190–1213, 2016.

[11] G. Lai, W. C. Chang, Y. Yang, and H. Liu, "Modeling Long- and Short-Term Temporal Patterns with Deep Neural Networks," arXiv: 1703.07015, 2017.

[12] A. Zimek, and E. Schubert, "Outlier Detection," Encyclopedia of Database Systems, Springer New York, pp. 1–5, 2019.

[13] T. Y. Kim, S. B. Cho, "Web traffic anomaly detection using C-LSTM neural networks," Expert Systems with Application, Vol. 106, pp. 66–76, 2018. DOI: 10.1016/j.eswa.2018.04.004

[14] Y. Jiang, C. Zeng, J. Xu, and T. Li, "Real time contextual collective anomaly detection over multiple data streams," Proceedings of the ODD, 2014. DOI: 10.1145/2656269.2656271

[15] L. Akoglu, H. Tong, and D. Koutra, "Graph based anomaly detection and description: a survey," Data Min Knowl Disc 29, 626–688, 2015. DOI: 10.1007/s10618-014-0365-y

[16] M.A. Hayes, and M.A. Capretz, "Contextual anomaly detection framework for big sensor data," Journal of Big Data Vol. 2, Issue 2, 2015. DOI: 10.1186/s40537-014-0011-y

[17] A.M. Kosek, "Contextual anomaly detection for cyber-physical security in Smart Grids based on an artificial neural network model," Joint Workshop on Cyber- Physical Security and Resilience in Smart Grids (CPSR-SG), pp. 1-6, 2016. DOI: 10.1109/CPSRSG.2016.7684103

Preprint: A Survey on Anomaly Detection for Technical Systems using LSTM Networks

[18] M. Munir, S. Erkel, A. Dengel, S. Ahmed, "Pattern-Based Contextual Anomaly Detection in HVAC Systems," Proceedings of 17th IEEE International Conference on Data Mining Workshops, IEEE Computer Society, pp. 1066–1073, 2017. DOI: 10.1109/ICDMW.2017.150

[19] S. Hochreiter, and J. Schmidhuber "Long short-term memory," Neural Computation, vol. 9, no. 8, pp. 1735-1780, 1997. DOI: 10.1162/neco.1997.9.8.1735

[20] B. Lindemann, T. Müller, H. Vietz, N. Jazdi and M. Weyrich, "A Survey on Long Short-Term Memory Networks for Time Series Prediction," 2020 14th CIRP Conference on Intelligent Computation in Manufacturing Engineering, Gulf of Naples, 2020, DOI: 10.13140/RG.2.2.36761.65129/1

[21] P. Malhotra, L. Vig, G. Shrof, and P. Agarwal, "Long Short Term Memory Networks for Anomaly Detection in Time Series," Proceedings of European Symposium on Artificial Neural Networks, Computational Intelligence and Machine Learning (ESANN), pp. 89-94, 2015.

[22] A. Taylor, S. Leblanc, and N. Japkowicz, "Anomaly Detection in Automobile Control Network Data with Long Short-Term Memory Networks," IEEE International Conference on Data Science and Advanced Analytics (DSAA), pp. 130–139, 2016. DOI: 10.1109/DSAA.2016.20

[23] T. Ergen, A. H. Mirza, and S. S. Kozat, "Unsupervised and Semi-supervised Anomaly Detection with LSTM Networks," arXiv:1710.09207, 2017.

[24] L. Bontemps, V. L. Cao, J. McDermott, and N. A. Le-Khac, "Collective Anomaly Detection based on Long Short Term Memory Recurrent Neural Networks," International Conference on Future Data and Security Engineering, pp. 141-152, 2016. DOI: 10.1007/978-3-319-48057-2_9

[25] M.-C. Lee, J.-C. Lin, and E.G. Gan, "ReRe: A Lightweight Real-Time Ready-to-Go Anomaly Detection Approach for Time Series," 2020 IEEE 4th Annual Computers, Software and Applications Conference (COMPSAC), Madrid, Spain, pp. 322-327, 2020. DOI: 10.1109/COMPSAC48688.2020.0-226

[26] J. Schmidhuber, "Deep learning in neural networks: an overview," Neural Netw, Vol. 61, pp. 85–117, 2015. DOI: 10.1016/j.neunet.2014.09.003

[27] C. Zhou, R. C. Paffenroth, "Anomaly Detection with Robust Deep Autoencoders," Proceedings of the 23rd ACM SIGKDD International Conference on Knowledge Discovery and Data Mining, New York, NY, pp. 665–674, 2017. DOI: 10.1145/3097983.3098052

[28] S. Naseer et al., "Enhanced Network Anomaly Detection Based on Deep Neural Networks," in IEEE Access, vol. 6, pp. 48231-48246, 2018. DOI: 10.1109/ACCESS.2018.2863036

[29] D. Park, Y. Hoshi, and C. Kemp, "A Multimodal Anomaly Detector for Robot-Assisted Feeding Using an LSTM-Based Variational Autoencoder," IEEE Robotics and Automation, Vol. 3, No. 3, 2018. DOI: 10.1109/LRA.2018.2801475

[30] B. Lindemann, N. Jazdi and M. Weyrich, "Anomaly detection and prediction in discrete manufacturing based on cooperative LSTM networks," 2020 IEEE 16th International Conference on Automation Science and Engineering (CASE), Hong Kong, Hong Kong, 2020, pp. 1003-1010, DOI: 10.1109/CASE48305.2020.9216855

[31] I. Sutskever, O. Vinyals, and Q. V. Le, "Sequence to sequence learning with neural networks," Advances in Neural Information Processing Systems, 27, 2014.

[32] T. Fernando, S. Denman, S. Sridharan, and C. Fookes, "Soft + Hardwired Attention: An LSTM Framework for Human Trajectory Prediction and Abnormal Event Detection," Neural Networks, Vol. 108, pp. 406- 478, 2017. DOI: 10.1016/j.neunet.2018.09.002

[33] G. Loganathan, J. Samarabandu, and X. Wang, "Sequence to Sequence Pattern Learning Algorithm for Real-Time Anomaly Detection in Network Traffic," IEEE Canadian Conference on Electrical & Computer Engineering (CCECE), pp. 1–4, 2018. DOI: 10.1109/CCECE.2018.8447597

[34] T. Kieu, B. Yang, C. Guo, and C. S. Jensen, "Outlier Detection for Time Series with Recurrent Autoencoder Ensembles," Proceedings of the 28th International Joint Conference on Artificial Intelligence (IJCAI), pp. 2725-2732, 2019. DOI: 10.24963/ijcai.2019/378

[35] Z. Li, J. Li, Y. Wang, and K. Wang, "A deep learning approach for anomaly detection based on SAE and LSTM in mechanical equipment," The International Journal of Advanced Manufacturing Technology, Vol. 103, pp. 499–510, 2019. DOI: 10.1007/s00170-019-03557-w

[36] B. Lindemann, F. Fesenmayr, N. Jazdi, and M. Weyrich, "Anomaly detection in discrete manufacturing using self-learning approaches," Procedia CIRP, vol. 79, pp. 313-318, 2019. DOI: 10.1016/j.procir.2019.02.073

[37] H. Zenati, C.S. Foo, B. Lecouat, G. Manek, and V. R. Chandrasekhar, "Efficient GAN-Based Anomaly Detection, " arXiv:1802.06222, 2018.

[38] T. Y. Kim, and S. B. Cho, "Web traffic anomaly detection using C-LSTM neural networks," Expert Systems with Applications, Vol. 106, pp. 66–76, 2018. DOI: 10.1016/j.eswa.2018.04.004

[39] S. Ding, A. Morozov, S. Vock, M. Weyrich, and K. Janschek, "Model-based Error Detection for Industrial Automation Systems using LSTM Networks," IMBSA: Model-Based Safety and Assessment, pp. 212-226, 2020. DOI: 10.1007/978-3-030-58920-2_14

[40] M.A. Prado-Romero, and A. Gago-Alonso, "Detecting contextual collective anomalies at a Glance," 23rd International Conference on Pattern Recognition (ICPR), pp. 2532-2537, 2016. DOI: 10.1109/ICPR.2016.7900017

[41] M.A. Haque and H. Mineno, "Contextual Outlier Detection in Sensor Data Using Minimum Spanning Tree Based Clustering," 2018 International Conference on Computer, Communication, Chemical, Material and Electronic Engineering (IC4ME2), pp. 1-4, 2018. DOI: 10.1109/IC4ME2.2018.8465643

[42] L. Zheng, Z. Li, J. Li, Z. Li, and J. Gao, "AddGraph: Anomaly Detection in Dynamic Graph Using Attention-based Temporal GCN," IJCAI, 2019.

[43] B. Maschler, and M. Weyrich, "Deep Transfer Learning for Industrial Automation," IEEE Industrial Electronics Magazine, in print, 2021. DOI: 10.1109/MIE.2020.3034884

[44] H. Tercan, A. Guajardo, and T. Meisen, "Industrial Transfer Learning: Boosting Machine Learning in Production," in 2019 IEEE 17th International Conference on Industrial Informatics (INDIN): Aalto University, Helsinki-Espoo, Finland, 22-25 July, 2019 : proceedings, Helsinki, Finland, pp. 274–279, 2019. DOI: 10.1109/INDIN41052.2019.8972099

[45] S. J. Pan and Q. Yang, "A Survey on Transfer Learning," IEEE Transactions on Knowledge and Data Engingeering, Vol. 22, No. 10, pp. 1345–1359, 2010. DOI: 10.1109/TKDE.2009.191

[46] C. Tan, F. Sun, T. Kong, W. Zhang, C. Yang, and C. Liu, "A Survey on Deep Transfer Learning," in Lecture Notes in Computer Science, Artificial Neural Networks and Machine Learning – ICANN 2018, V. Kůrková, Y. Manolopoulos, B. Hammer, L. Iliadis, and I. Maglogiannis, Eds., Cham: Springer International Publishing, pp. 270–279, 2018. DOI: 10.1007/978-3-030-01424-7_27

[47] P. Liang, H.-D. Yang, W.-S. Chen, S.-Y. Xiao, and Z.-Z. Lan, "Transfer learning for aluminium extrusion electricity consumption anomaly detection via deep neural networks," International Journal of Computer Integrated Manufacturing, Vol. 31, No. 4-5, pp. 396–405, 2018. DOI: 10.1080/0951192X.2017.1363410

[48] M. Canizo, I. Triguero, A. Conde, and E. Onieva, "Multi-head CNN–RNN for multi-time series anomaly detection: An industrial case study," Neurocomputing, vol. 363, pp. 246–260, 2019. DOI: 10.1016/j.neucom.2019.07.034

[49] R.-J. Hsieh, J. Chou, and C.-H. Ho, "Unsupervised Online Anomaly Detection on Multivariate Sensing Time Series Data for Smart Manufacturing," in 2019 IEEE 12th Conference on Service-Oriented Computing and Applications (SOCA), Kaohsiung, Taiwan, pp. 90–97, 2019. DOI: 10.1109/SOCA.2019.00021

[50] S. Tariq, S. Lee, and S. S. Woo, "CANTransfer: transfer learning based intrusion detection on a controller area network using convolutional LSTM network," in Proceedings of the 35th Annual ACM Symposium on Applied Computing (SAC), Brno, Czech Republic, pp. 1048–1055, 2020. DOI: 10.1145/3341105.3373868

[51] L. Fei-Fei, R. Fergus, and P. Perona, "One-shot learning of object categories," IEEE Transactions on Pattern Analysis and Machine Intelligence, Vol. 28, No. 4, pp. 594–611, 2006. DOI: 10.1109/TPAMI.2006.79

[52] B. Maschler, T.T.H. Pham and M. Weyrich, "Regularization-based Continual Learning for Anomaly Detection in Discrete Manufacturing," 2021 54th CIRP Conference on Manufacturing Systems, Athens, 2021. arXiv:2101.00509

[53] J. Kirkpatrick, R. Pascanu, N. Rabinowitz, J. Veness, G. Desjardins et al., „Overcoming catastrophic forgetting in neural networks," Proceedings of the National Academy of Sciences, Vol. 114, No. 13, pp. 3521-3526. DOI: 10.1073/pnas.1611835114

[54] J. Schwarz, J. Luketina, W.M. Czarnecki1, A. Grabska-Barwinska1, Y.W. The et al., "Progress & Compress: A scalable framework for continual learning," Proceedings of Machine Learning Research, Vol. 80, pp. 4528-4537, 2018. arXiv:1805.06370